\documentclass[letterpaper, 10 pt, conference]{ieeeconf}  

\IEEEoverridecommandlockouts                              

\usepackage{url}
\usepackage{graphicx} 
\usepackage{amsmath} 
\usepackage{amssymb}  

\usepackage{algorithm,algpseudocode}
\usepackage{algorithm}
\usepackage{xspace}
\usepackage{xcolor}

\makeatletter
\let\NAT@parse\undefined
\makeatother
\usepackage{hyperref}
\hypersetup{linkbordercolor=green}

\DeclareMathOperator*{\argmax}{arg\,max}
\graphicspath{ {img/} }

\algrenewcommand{\algorithmiccomment}[1]{\hfill // \textit{#1}}

\title{\LARGE \bf
An On-Line POMDP Solver for Continuous Observation Spaces
}

\author{Marcus Hoerger$^{1}$ and Hanna Kurniawati$^{1}$
\thanks{*This work is partially funded by the ANU Futures Scheme QCE20102.}
\thanks{$^{1}$College of Engineering \& Computer Science,
        The Australian National University, 7500 Canberra, Australia
        {\tt\small \{marcus.hoerger, hanna.kurniawati\}@anu.edu.au}}%
}

\newcommand{\ie}{\textrm{i.e.}\xspace}

\newcommand{\ccite}[1]{~\cite{#1}}

\newcommand{\sref}{Section~\ref}

\newcommand{\aref}[1]{Algorithm~\ref{#1}}
\newcommand{\eref}[1]{eq.(\ref{#1})}
\newcommand{\fref}[1]{Fig.~\ref{#1}}

\newcommand{\tref}[1]{Table~\ref{#1}}

\newcounter{comment}

\newcommand{\transF}{\ensuremath{T}\xspace}
\newcommand{\transFComp}{\ensuremath{T(s, a, s')}\xspace}
\newcommand{\obsF}{\ensuremath{Z}\xspace}

\newcommand{\rewFunc}{\ensuremath{R}\xspace}

\newcommand{\policy}{\ensuremath{\pi}\xspace}

\newcommand{\optPol}{\ensuremath{\pi^*}\xspace}

\newcommand{\stSpace}{\ensuremath{\mathcal{S}}\xspace}

\newcommand{\st}{\ensuremath{s}\xspace}
\newcommand{\stp}{\ensuremath{s'}\xspace}
\newcommand{\actSpace}{\ensuremath{\mathcal{A}}\xspace}
\newcommand{\act}{\ensuremath{a}\xspace}
\newcommand{\obsSpace}{\ensuremath{\mathcal{O}}\xspace}
\newcommand{\obs}{\ensuremath{o}\xspace}

\newcommand{\belSpace}{\ensuremath{\mathcal{B}}\xspace}
\newcommand{\bel}{\ensuremath{b}\xspace}

\newcommand{\solverShort}{LABECOP\xspace}
\newcommand{\solver}{Lazy Belief Extraction for Continuous Observation POMDPs\xspace}

\newcommand{\episode}{\ensuremath{h}\xspace}
\newcommand{\episodeSet}{\ensuremath{H}\xspace}
\newcommand{\del}[1]{\mathrm{d}{#1}}

\algnewcommand{\IIf}[1]{\State\algorithmicif\ #1\ \algorithmicthen}
\algnewcommand{\EndIIf}{\unskip\ \algorithmicend\ \algorithmicif}
\algnewcommand{\IfThenElse}[3]{
  \State \algorithmicif\ #1\ \algorithmicthen\ #2\ \algorithmicelse\ #3}

\begin{document}

\maketitle
\thispagestyle{empty}
\pagestyle{empty}

\begin{abstract}

Planning under partial obervability is essential for autonomous robots. A principled way to address such planning problems is the Partially Observable Markov Decision Process (POMDP). Although solving POMDPs is computationally intractable, substantial advancements have been achieved in developing approximate POMDP solvers in the past two decades.
However, computing robust solutions for problems with continuous observation spaces remains challenging. Most on-line solvers rely on discretising the observation space or artificially limiting the number of observations that are considered during planning to compute tractable policies. In this paper we propose a new on-line POMDP solver, called \solver (\solverShort), that combines methods from Monte-Carlo-Tree-Search and particle filtering to construct a policy reprentation which doesn't require discretised observation spaces and avoids limiting the number of observations considered during planning. Experiments on three different problems involving continuous observation spaces indicate that \solverShort performs similar or better than state-of-the-art POMDP solvers.

\end{abstract}

\section{INTRODUCTION}
Planning under partial observability is both challenging and essential for autonomous robots. The Partially Observable Markov Decision Processes (POMDP)\ccite{Sondik:71} is a mathematically principled way to solve such planning problems. By lifting the planning problem from the robot's state space to its \textit{belief space}, \ie the set of all probability distributions over the state space, POMDPs enable autonomous robots to systematically reason about different strategies to achieve a given task while having only access to noisy or incomplete observations. 
Although solving a POMDP exactly is computationally intractable\ccite{papadimitriou1987complexity}, the past two decades have seen tremendous progress in developing approximately optimal solvers that trade optimality for computational tractability, making them applicable to various robotic planning problems\ccite{bai2012unmanned,horowitz2013interactive,hsiao2007grasping,hoerger2019candy}.

Despite the mentioned advances, solving POMDP problems with continuous observation space remains a challenge. Existing on-line POMDP solvers require some kind of discretisation of the observation space, either explicitly or implicitly.  

However, discretising the observation space can be difficult, particularly for high-dimensional observation spaces, since it requires a good trade-off between accuracy and planning efficiency. Finer discretisations of the observation space lead to a substantial increase in computation time, whereas coarser discretisations can impair the quality of the resulting policies. 
More recently,\ccite{sunberg2018online} proposed to avoid discretising the observation space by sampling and keeping sampled observations only occasionally. However, this strategy can be sensitive to the rate at which observations are sampled and kept.

In this paper we propose a new on-line POMDP solver, called \solver (\solverShort) to alleviate the above issues.
\solverShort avoids any form of observation space discretisation by realising that a set of sampled episodes ---that is, sequences of state--action--observation--reward quadruples--- is sufficient to represent many different belief sequences and that beliefs, along with their action-values and action-selection strategy, can be estimated using this set via an episode re-weighting method inspired by particle filters. 
During planning, \solverShort extracts sequences of beliefs and their corresponding action-values and action-selection strategy lazily, in a sense that they are only extracted for the action-observation sequence of the currently sampled episode.
This allows \solverShort to consider \textit{every} sampled observation and therefore \textit{every} sequence of beliefs it encounters during planning, without having to discretise the observation space.

Experimental results on three test scenarios indicate that \solverShort is more effective in computing robust policies for problems with continuous observation spaces compared to state-of-the-art methods.

\section{BACKGROUND AND RELATED WORK}
\subsection{Partially Observable Markov Decision Process (POMDP)}
Formally a POMDP is a tuple $<\stSpace, \actSpace, \obsSpace, \transF, \obsF, \rewFunc, \gamma>$, where \stSpace, \actSpace and \obsSpace are the state, action and observation spaces of the robot. \transF and \obsF model the uncertainty in the effect of taking actions and receiving observations as conditional probability functions $\transFComp = p(\stp | \st, \act)$ and $Z(\stp, \act, \obs) = p(\obs | \stp, \act)$, where $\st, \stp \in \stSpace$, $\act \in \actSpace$ and $\obs \in \obsSpace$. $R(\st, \act)$ models the reward the robot receives when performing action $\act$ from $\st$ and $0 < \gamma < 1$ is a discount factor. Due to uncertainties in the effect of performing actions and receiving observations, the true state of the robot is only partially observable. Hence, instead of planning with respect to states, the robot plans with respect to probability distributions $\bel \in \belSpace$ over the state space, called beliefs, where $\belSpace$ is the set of all probability distributions over \stSpace.
The solution of a POMDP is an optimal policy \optPol, a mapping from beliefs to actions $\optPol: \bel \mapsto \act$ such that the robot maximises the expected discounted future reward when following \optPol. Once \optPol has been computed, it can be used as a feedback-controller: Given the current belief $\bel$, the robot performs $\optPol(\bel)$, receives an observation $\obs \in \obsSpace$ and updates its belief according to $\bel' = \tau(\bel, \act, \obs)$, where $\tau$ is the Bayesian belief update function. The value achieved by a policy \policy at a particular belief \bel can be expressed as
\begin{equation}\label{eq:policy_value}
    V_{\policy}(\bel) = R(\bel, \policy(\bel)) + \gamma \int_{\obs \in \obsSpace} Z(\bel, \policy(\bel), \obs) V_{\policy}(\tau(\bel, \policy(\bel), \obs)) \del o
\end{equation}

where $R(\bel, \act) = \int_{\st \in \stSpace} R(\st, \act) \bel(s) \del s$ and $Z(\bel, \act, \obs) = \int_{\stp \in \stSpace} Z(\stp, \act, \obs) \int_{\st \in \stSpace}\transFComp\bel(\st)\del \st\del\stp$. The optimal policy \optPol is then the policy that satisfies $\optPol(\bel) = \argmax_{\policy} V_{\policy}(\bel)$.

\subsection{Related On-line POMDP Solvers}
POMDP solvers have seen tremendous advances in the past two decades. Key to their success is the use of sampling to trade optimality for computational tractability. These solvers can be broadly classified into off-line and on-line solvers. While off-line solvers\ccite{Kurniawati08sarsop:efficient, Pin03:Point, Smi05:Point} compute an approximate-optimal policy for a sampled representation of the belief space, on-line solvers interleave planning and execution by computing a policy for the current belief only. Various on-line solvers have been proposed for continuous state spaces\ccite{silver2010monte,somani2013despot,kurniawati2016online}, large or continuous actions spaces\ccite{seiler2015online,Wang2018}, expensive transition dynamics\ccite{hoergerISRR2019} and long planning horizons\ccite{agha2011firm,He2010:Puma,kurniawati2011motion}. To compute a policy for the current belief, these solvers typically perform forward-search in a lookahead-tree to evaluate sequences of actions starting from the current belief. For instance POMCP\ccite{silver2010monte} and ABT\ccite{kurniawati2016online} use Monte-Carlo-Tree-Search by sampling many sequences of actions. This allows them to quickly focus the search on the most promising parts of the lookahead-tree. DESPOT\ccite{somani2013despot} uses a combination of Monte-Carlo sampling, heuristic search and branch-and-bound pruning to construct a sparse representation of the lookahead-tree.

For continuous observation spaces most of the aforementioned methods require some form of discretisation of the observation space to trade the branching factor of the lookahead-tree with accuracy. More recently,\ccite{garg2019despot} extended DESPOT to handle large discrete observation spaces. While this approach allows for a finer discretisation of the observation space, it can't handle purely continuous observation spaces. Another approach is POMCPOW\ccite{sunberg2018online} and extension of POMCP to handle continuous action and observation spaces. POMCPOW uses Progressive-Widening\ccite{couetoux2011continuous} to slowly add sampled observations to the lookahead-tree as planning progresses. While this approach avoids discretising the observation space, the rate at which sampled observations are added to the tree must be carefully chosen in order to keep the branching factor of the search tree manageable. 

\section{LAZY BELIEF EXTRACTION FOR CONTINUOUS OBSERVATION POMDPS (LABECOP)}
\subsection{Overview}\label{ssec:methodOverview}

\begin{figure*}
\vspace{-0.5cm}
\centering
\small
\begin{tabular}{c@{\hskip15pt}c@{\hskip15pt}c@{\hskip15pt}c@{\hskip15pt}c}
\includegraphics[width=0.18\textwidth]{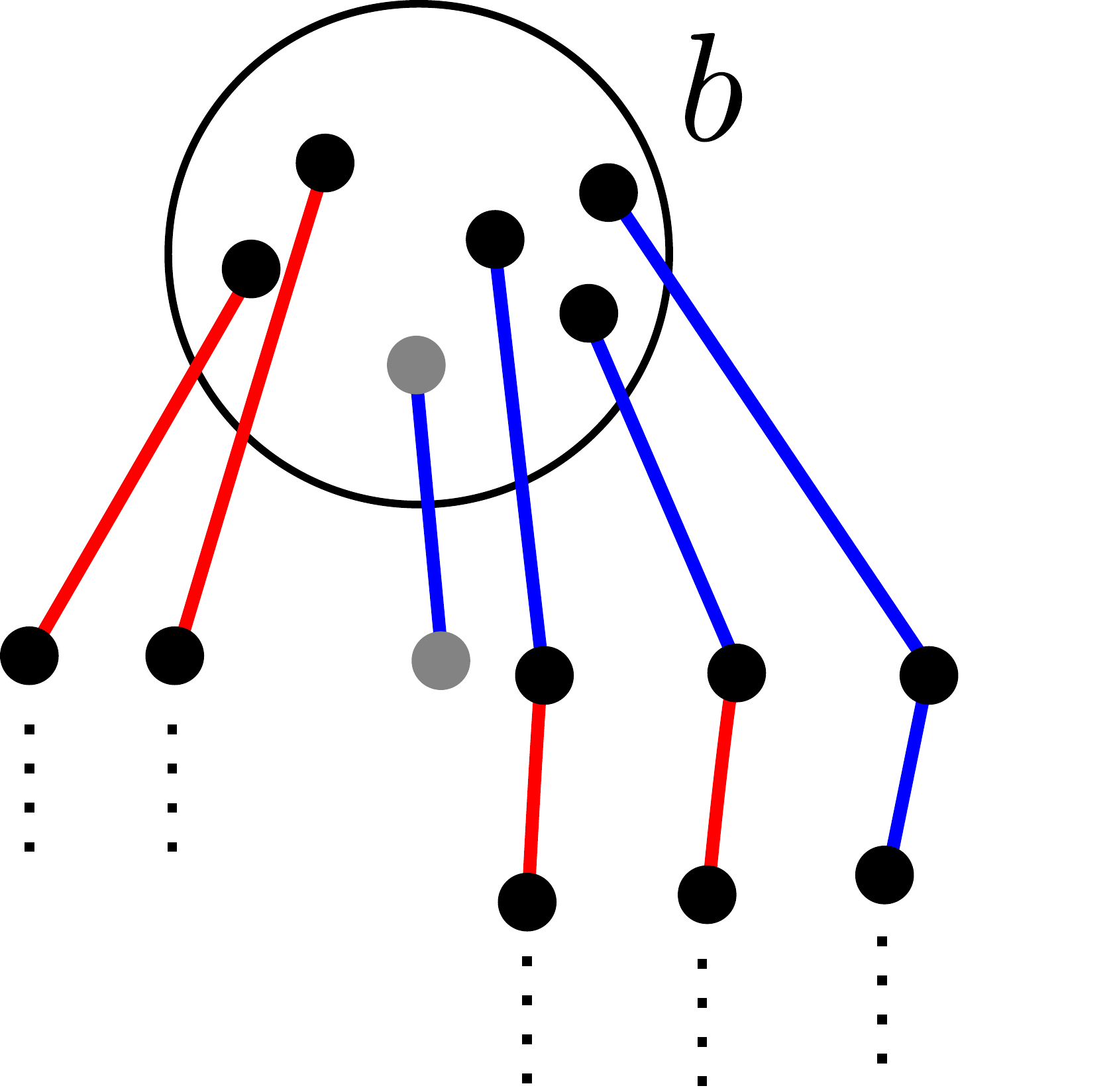} &
\includegraphics[width=0.18\textwidth]{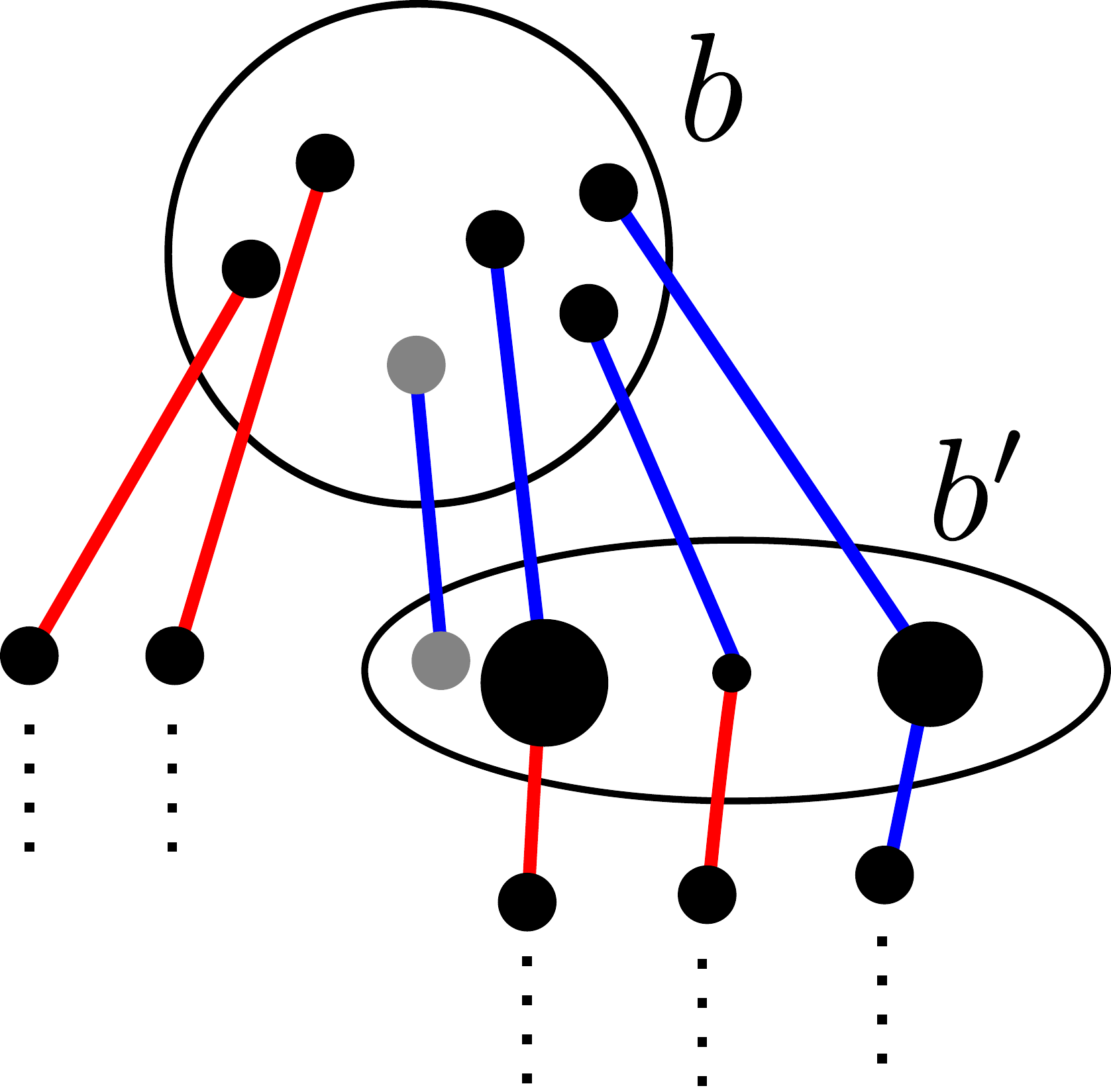} &
\includegraphics[width=0.18\textwidth]{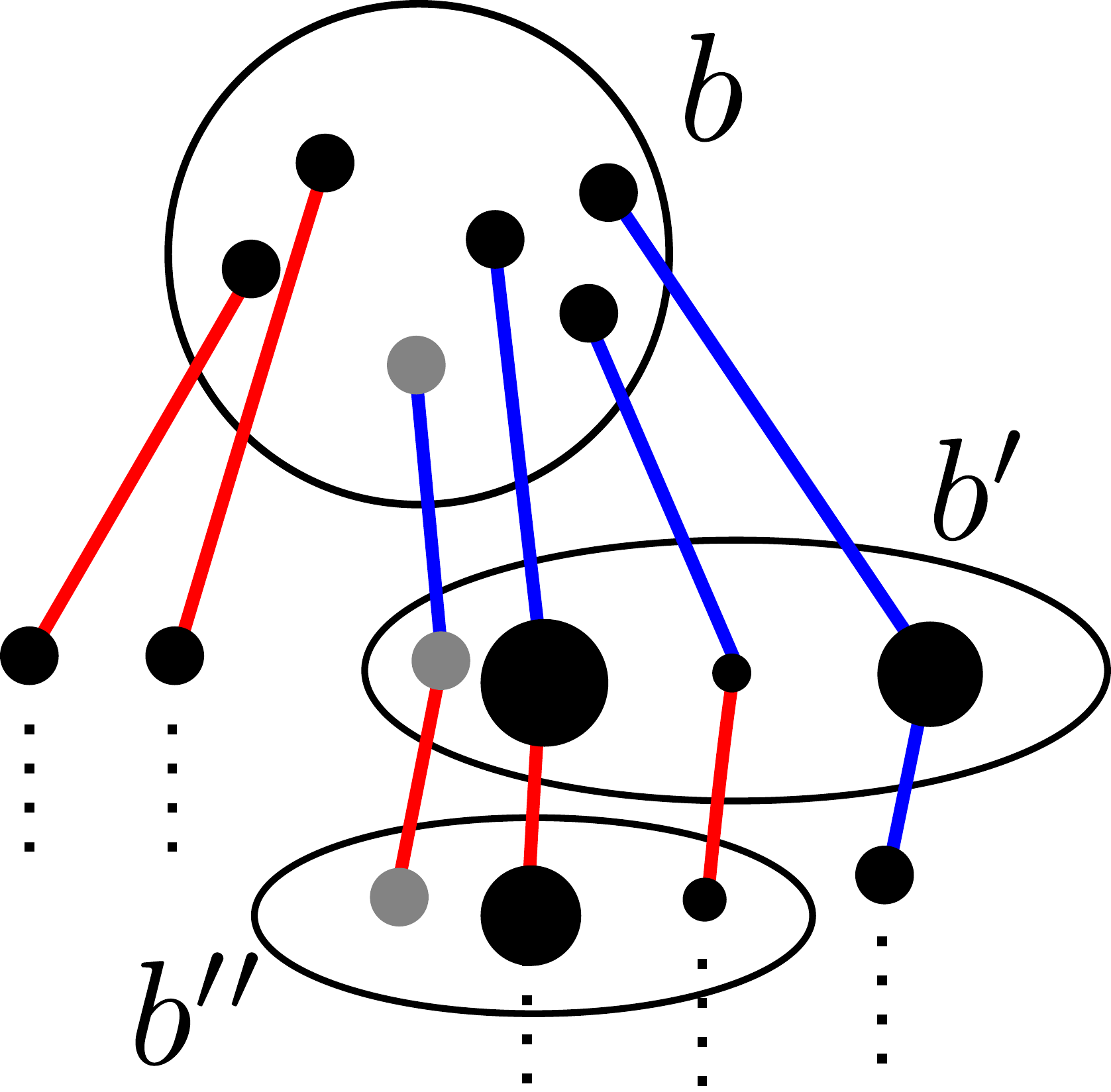} &
\includegraphics[width=0.18\textwidth]{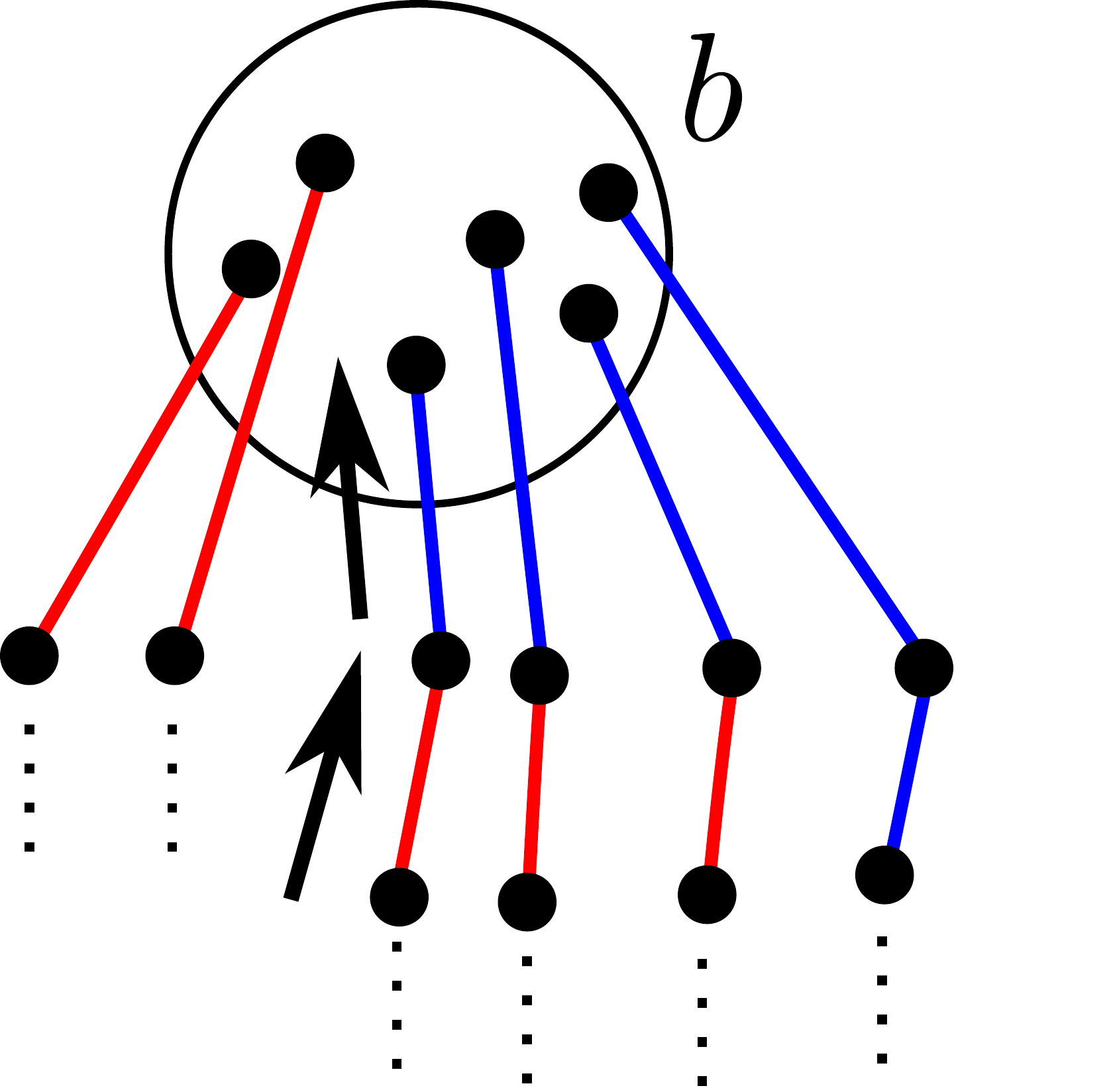} \\
(a) & (b) & (c) & (d)
\end{tabular}
\caption{Illustration of the episode sampling process for a POMDP with two actions $\act_1$ and $\act_2$. (a) Given a set of episodes $\episodeSet_{(\bel)}$ (shown as black dots connected by action edges, where the dots represent states and the edges represent actions -- blue for $\act_1$ and red for $\act_2$) that start from the current belief \bel (shown as circle) we first sample a state (shown as grey dot) from the current belief, select an action according to \eref{eq:ucb_with_weights} ($\act_1$ in this case), sample a next state from the transition function (second grey dot) and an observation $\obs$. (b) We compute a weight (represented by the size of the black dots) for each state associated with the second quadruples of the episodes in $H_{(\bel, \act_1)}$ (black oval), based on the selected action $\act_1$, the sampled observation and the weights of the parent states, giving us a approximation of $\bel' = \tau(\bel, \act_1, \obs)$. (c) From $\bel'$, we select a new action according to \eref{eq:ucb_with_weights} sample a next state and observation and continue from the set $H_{(\bel', \act_1)}$ (second black oval). (d) After sampling the episode, we perform Bellman backup to compute its values, add the episode to $\episodeSet_{(\bel)}$ and discard the previously computed weights.}
\label{f:intuition}
\vspace{-0.5cm}
\end{figure*}

\solverShort is an anytime on-line POMDP solver designed for POMDP problems with continuous state and observation spaces. \solverShort incrementally samples a set of \textit{episodes} $\episodeSet_{(\bel)}$, \ie sequences of state-action-observation-reward quadruples starting from the current belief $\bel\in\belSpace$ to compute an approximation of the optimal policy for \bel. 

Key to \solverShort is the realisation that the episodes in $\episodeSet_{(\bel)}$ provide sufficient information to encode many different sequences of beliefs starting from \bel and that these beliefs, along with their approximated action values and corresponding action-selection strategy, can be extracted from $\episodeSet_{(\bel)}$ on-the-fly during planning, without having to maintain a lookahead-tree. Both components -- sequence of visited beliefs and the action-selection strategy -- are extracted lazily from $\episodeSet_{(\bel)}$ in a sense that we only extract them for the currently sampled episode. This enables \solverShort to consider \textit{every} sampled observation (and therefore \textit{every} encountered belief) during planning which helps \solverShort in discovering good policies that rely on differentiating between many sequences of beliefs. \fref{f:intuition} illustrates the core idea behind \solverShort for a POMDP with two actions. 

Whenever we sample a new episode \episode, we iteratively select and re-weight subsets of $\episodeSet_{(\bel)}$, based on the actions and sampled observations associated to \episode to extract the sequence of beliefs visited by \episode. Moreover, each quadruple of the episodes in $\episodeSet_{(\bel)}$ has an associated value that is computed only once per episode. We use these values to approximate the action values $Q(\bel, \act)$ (\ie the expected value of executing $\act\in\actSpace$ from \bel and continuing optimally afterwards) at the extracted beliefs via weighted sums of the episode values. The approximated action values are then used to derive an action-selection strategy. The methods that we use to extract sequences of beliefs and derive an action-selection strategy are described in \sref{ssec:extractingBeliefs} and \sref{ssec:actionSelectionStrategy} respectively.

To sample a new episode, we first sample an initial state $\st\in\stSpace$ from the current belief $\bel$ and select an action $\act\in\actSpace$ according to the strategy described in \sref{ssec:actionSelectionStrategy}. We then sample a next state $\stp\in\stSpace$, an observation $\obs\in\obsSpace$ and the immediate reward $r = R(\st, \act)$ by calling a generative back-box model $(\stp, \obs, r) = G(\st, \act)$ and add the quadruple $(\st, \act, \obs, r)$ to the episode. Given the selected action \act and perceived observation \obs, we can now extract an approximation of next belief $\bel' = \tau(\bel, \act, \obs)$ from $\episodeSet_{(\bel)}$ as described in \sref{ssec:extractingBeliefs} and repeat the above steps from \stp and $\bel'$. Upon completion, we add the newly sampled episode to $\episodeSet_{(\bel)}$. An overview of the episode sampling process is shown in \aref{alg:sampleEpisode}.

Once the planning time for the current step is over, \solverShort selects an action from the current belief \bel according to
\begin{equation}\label{eq:solver_policy}
\policy(\bel) = \argmax_{\act\in\actSpace} Q(\bel, \act)
\end{equation} 

After executing $\policy(\bel)$ and perceiving an observation, we update the belief using a SIR particle filter\ccite{arulampalam2002tutorial} and continue planning from the updated belief. This process repeats until a maximum number of planning steps is reached, or the agent enters a terminal state (we assume that we know when the agent enters a terminal state).

\subsection{Extracting a sequence of beliefs for a new episode}\label{ssec:extractingBeliefs}
After we have sampled an initial state, selected an action \act from the current belief \bel and perceived an observation \obs, we now have to obtain an approximation of the belief $\bel'=\tau(\bel, \act, \obs)$ to select the next action. As mentioned earlier, we can extract an approximation of $\bel'$ directly from $\episodeSet_{(\bel)}$. This is done as follows:
 
For every episode $\episode\in\episodeSet_{(\bel)}$, the state $\episode_{(1)}.\st$ associated to the first quadruple (we denote the $i$-th quadruple of \episode as $\episode_{(i)}$ and its associated state as $\episode_{(i)}.\st$) has an initial weight $w(\episode_{(1)}.\st)$ corresponding to the weight of the belief particle \episode starts from. If the current belief is represented by a set of unweighted particles, we set $w(\episode_{(1)}.\st) = 1$. 
Now, suppose $H_{(\bel, \act)}$ is the set of all episodes in $\episodeSet_{(\bel)}$ that contain action $\act$ in their first quadruple. For every episode $\episode\in\episodeSet_{(\bel, \act)}$ we recursively assign a weight to the state associated to the second quadruple of \episode according to $w(\episode_{(2)}.\st) = w(\episode_{(1)}.\st)Z(\episode_{(2)}.\st, \act, \obs) / \eta$, where $\eta$ is a normalisation constant. In other words, we compute the relative likelihood of "seeing" the sampled observation \obs given that we ended up in $\episode_{(2)}.\st$ after executing \act from $\episode_{(1)}.\st$. Thus, the collection of all states associated to the second quadruples of the episodes in $H_{(\bel, \act)}$ along with their computed weights forms a sampled approximation of $\bel'$. This is similar to a particle filter. 

Once we select the next action $\act'$ from $\bel'$ we set $\episodeSet_{(\bel')} = \episodeSet_{(\bel, \act)}$ and iteratively compute the weights $w(\episode_{(3)}.\st)$ for every $\episode\in\episodeSet_{(\bel',\act')}\subseteq\episodeSet_{(\bel')}$, given $w(\episode_{(2)}.\st)$, the selected action $\act'$ and the next observation. By repeating this process as we continue sampling the episode, we obtain a sequence of sampled approximations of the beliefs visited by the episode. Note that these weights are re-computed every time we sample a new episode since each sampled episode visits a different sequence of beliefs. \aref{alg:reweightEpisodes} provides an overview of this belief extraction process.

Intuitively, by extracting approximated beliefs using the weighting method above, we obtain richer belief approximations for action sequences that are visited more often, which helps \solverShort to evaluate frequently visited action sequences more accurately. On the other hand, extracting beliefs becomes more expensive as planning progresses, since we have to re-compute the weights for a growing set of episodes. Additionally, compared to many on-line POMDP solvers that only require a generative black-box model for planning, our method explicitly requires us to evaluate the observation function $Z(\st, \act, \obs)$ which might not be readily available. The first issue could be mitigated by numerical fitting, \ie weighting only a small subset of the states and then perform interpolation, similarly to\ccite{li2016numerical}. The second issue could be mitigated by using an approximation for $Z(\st, \act, \obs)$ which is often easy to obtain. We are planning to explore both options in future works.

\subsection{Action-selection strategy at the extracted beliefs}\label{ssec:actionSelectionStrategy}
To select actions from an approximated belief $\bel$ of depth $d$ obtained with the method above, \solverShort uses Upper Confidence Bounds1 (UCB1)\ccite{Aue02:Finite}, one of the most widely used strategies in on-line POMDP planning. UCB1 requires two quantities to select an action: Belief-dependent visitation counts $N(\bel)$ and $N(\bel, \act)$, \ie the number of times \bel has been expanded and the number of times each action has been used to expand \bel, as well as an estimate $\widehat{Q}(\bel, \act)$ of the $Q$-values. Since we never visit the same belief twice, except for the current belief, we slightly modify UCB1:

Let $W(\bel) = \sum_{\episode\in\episodeSet_{(\bel)}} w(\episode_{(d)}.\st)$ and $W(\bel, \act) = \sum_{\episode\in\episodeSet_{(\bel, \act)}} w(\episode_{(d)}.\st)$. $W(\bel)$ can be seen as the weighted number of times $\bel$ has been visited and $W(\bel, \act)$ as the weighted number of times $\act$ has been used to expand $\bel$. However, we cannot use $W(\bel)$ and $W(\bel,\act)$ in lieu of $N(\bel)$ and $N(\bel,\act)$ since $W(\bel)$ can be smaller than 1, for which UCB1 is undefined. To resolve this we replace $N(\bel)$ with $N_+(\bel)$, which is the number of episodes in $\episodeSet_{(\bel)}$ for which $w(\episode_{(d)}.\st)$ is greater than $0$, and replace $N(\bel, \act)$ with a scaled action weight $\widetilde{W}(\bel, \act) = \frac{W(\bel, \act)}{W(\bel)}N_+(\bel)$.

The last ingredient we need for UCB1 are the $Q$-value estimates. It turns out that, similarly to the beliefs, we can extract those values on-the-fly from $\episodeSet_{(\bel)}$ for the currently sampled episode. For a belief of depth $d$, \solverShort approximates $Q(\bel, \act)$ according to
\begin{equation}\label{eq:q_approx}
\widehat{Q}(\bel, \act) = \frac{1}{W(\bel, \act)}\sum_{\episode\in\episodeSet_{(\bel, \act)}}w(\episode_{(d)}.\st)V(\episode_{(d)})
\end{equation}
 
where $V(\episode_{(d)})$ is the value of the $d$-th quadruple of episode \episode, \ie an approximation of the expected value of executing $\episode_{(d)}.\act$ from $\episode_{(d)}.\st$ and continuing optimally after seeing $\episode_{(d)}.\obs$. These values are computed only once per episode via Bellman backup. Using $N_+(\bel)$, $\widetilde{W}(\bel, \act)$, and $\widehat{Q}(\bel, \act)$ we can now apply UCB1, \ie we select an action according to
\begin{equation}\label{eq:ucb_with_weights}
a = \argmax_{\act\in\actSpace}\left( \widehat{Q}(\bel, \act) + c\sqrt{\frac{\log(N_{+})}{\widetilde{W}(\bel,\act)}}\right)
\end{equation}

Note that during the episode sampling process we might end up in an approximated belief for which $\widetilde{W}(\bel, \act)$ is zero for one or more actions. If this is the case we treat these actions as "unvisited" and use a rollout strategy that selects one of these actions uniformly at random. Additionally, we compute a problem-dependent heuristic estimate for $V(\episode_{(k)})$ given $\episode_{(k)}.\st$ and the randomly selected action, where $\episode_{(k)}$ is the last quadruple of \episode. Finally we backup the episode along the sequence of visited beliefs to obtain $V(\episode_{(i)})$ for each quadruple of the episode and add the episode to $\episodeSet_{(\bel)}$. An overview of the action-selection process is shown in \aref{alg:selectUCBAction}.

\begin{algorithm}
\caption{SampleEpisode(Current belief \bel, Set of episodes $\episodeSet_{(\bel)})$}\label{alg:sampleEpisode}
\begin{algorithmic}[1]  
  \State \episode = init episode; $i = 1$; $\st \sim \bel$; unvisitedAction = False
  \While{unvisitedAction is False and \st not terminal}
     \State (\act, unvisitedAction) = \texttt{SelectAction}(\bel, $\episodeSet_{(\bel)}$)\Comment{\aref{alg:selectUCBAction}}
     \State $(\stp, \obs, r) = G(\st, \act)$\Comment{Generative black-box model}     
     \State Insert $(\st, \act, \obs, r)$ to \episode
     \State $\episodeSet_{(\bel, \act)} = \{\episode\in\episodeSet_{(\bel)} | \episode_{(i)}.\act = \act\}$     
     \State $\bel' = $\ \texttt{ExtractBelief}(\bel, $\episodeSet_{(\bel, \act)}$, \act, \obs, $i$)\Comment{\aref{alg:reweightEpisodes}}
     \State $\episodeSet_{(\bel')} = \episodeSet_{(\bel, \act)}$; $\st = \stp$; $\bel = \bel'$; $i = i + 1$
  \EndWhile
  \If{unvisitedAction is True} 
    \State $V(h_{(i)}) =$ calculateHeuristic($\st$, \episode) 
  \EndIf
  \State insert $(\st, -, -, 0)$ to \episode  
  \State backupEpisode(\episode, $V(h_{(i)})$)
  \State \Return \episode   
\end{algorithmic}
\end{algorithm}
\begin{algorithm}
\caption{ExtractBelief(Belief \bel, Set of episodes $\episodeSet$, Action \act, Observation \obs, Index $i$)}\label{alg:reweightEpisodes}
\begin{algorithmic}[1]
  \State $\bel' = \varnothing$;
    \For{$\episode\in\episodeSet$}
     \State $w(\episode_{(i+1)}.\st) = \bel(\episode_{(i)}.\st)Z(\episode_{(i+1)}.\st, \act, \obs) / \eta$\Comment{$\bel(\episode_{(i)}.\st) = w(\episode_{(i)}.\st)$}
     \State $\bel' = \bel' \cup \{(\episode_{(i+1)}.\st, w(\episode_{(i+1)}.\st))\}$     
  \EndFor   
  \State \Return $\bel'$
\end{algorithmic}
\end{algorithm}
\begin{algorithm}
\caption{SelectAction(Belief $\bel$, Set of Episodes $\episodeSet_{(\bel)}$, Index i}\label{alg:selectUCBAction}
\begin{algorithmic}[1]
  \State $N_+(\bel) =$\ \#episodes in $\episodeSet_{(\bel)}$ for which $\bel(\episode_{(i)}.\st)>0$\Comment{$\bel(\episode_{(i)}.\st)=w(\bel(\episode_{(i)}.\st))$}
  \State Calculate $W(\bel)$ and $W(\bel, \act)$, $\widetilde{W}(\bel, \act)$, $\widehat{Q}(\bel, \act)$ for each $\act\in\actSpace$
  \State $A_{unvisited} = \{\act \in \actSpace\ |\ \widetilde{W}(\bel, \act) = 0\}$ 
  \If{$\left |A_{unvisited} \right | > 0$}
    \State \Return (random action from $A_{unvisited}$, True)
  \EndIf
  \State $\act = \argmax_{\act\in\actSpace}\left( \widehat{Q}(\bel, \act) + c\sqrt{\frac{\log(N_{+})}{\widetilde{W}(\bel,\act)}}\right)$
  \State \Return (\act, False)  
\end{algorithmic}
\end{algorithm}

\section{EXPERIMENTS AND RESULTS}
To evaluate \solverShort, we tested and compared it with POMCPOW, POMCP, ABT and DESPOT on tree POMDP problems with continuous observation spaces, including an agent operating in a light-dark environment, a navigation problem with environment uncertainty and a motion-planning problem under uncertainty problem for a 4DOFs torque-controlled manipulator. The problem scenarios are detailed below. 

\subsection{Problem Scenarios}\label{ssec:problem_scenarios}
\subsubsection{LightDark1D}
The LightDark1D scenario is a simple benchmark problem introduced in\ccite{sunberg2018online} in which an agent navigates inside a one-dimensional discrete domain. The state space of the agent is the set of integers and it has access to four actions $\actSpace = \{-10, -1, 0, 1, 10\}$ to navigate deterministically inside the domain. Executing $a=0$ results in a terminal state and is rewarded by 100 at $\st=0$, but penalised by -100 at every other state. Additionally, the agent receives a penalty of $-1$ for every step it takes. Initially the agent is uncertain about its exact location but it receives  noisy one-dimensional continuous observations regarding its location that are more accurate in the vincinity of $\st=10$. The discount factor is 0.95. More details and an illustration of the problem can be found in\ccite{sunberg2018online}.

\subsubsection{Passage}
A robot equipped with a noisy and expensive range sensor operates on a 2D-plane populated by three obstacles and a border region. The robot must navigate from a known initial state to a goal area while avoiding collisions with the obstacles or the border region. The robot has perfect information regarding its own position, the border region as well as the goal area, but the locations of the obstacles are uncertain. The robot has access to 5 actions $\actSpace = \{FORWARD, LEFT, RIGHT, SCAN\}$, where the first three actions move the robot deterministically in the respective direction. The \textit{SCAN} action activates the range sensor and produces a three-dimensional continuous observation, consisting of the euclidean distances of the robot to the three obstacles. These observations become more precises as the robot navigates closer to an obstacle. Note that the robot only receives an observation when it executes the \textit{SCAN} action. The problem terminates upon colliding with the environment, which is penalised by -1,000 or reaching the goal area, which is rewarded by 1,000. Activating the range sensor results in a penalty of -50 and every step incurs a small penalty of -1. Here the discount factor is 0.95. \fref{f:problemScenarios}(a) illustrates the scenario, where the red square represents the robot, the light-grey and green areas represent the border region and the goal area respectively and the white squares are the obstacles. The shaded blue areas represent the initial (uniform) distributions of the obstacles.

\subsubsection{4DOF-Factory}
\begin{figure}
\centering
\begin{tabular}{c@{\hskip20pt}c}
\includegraphics[width=0.2\textwidth]{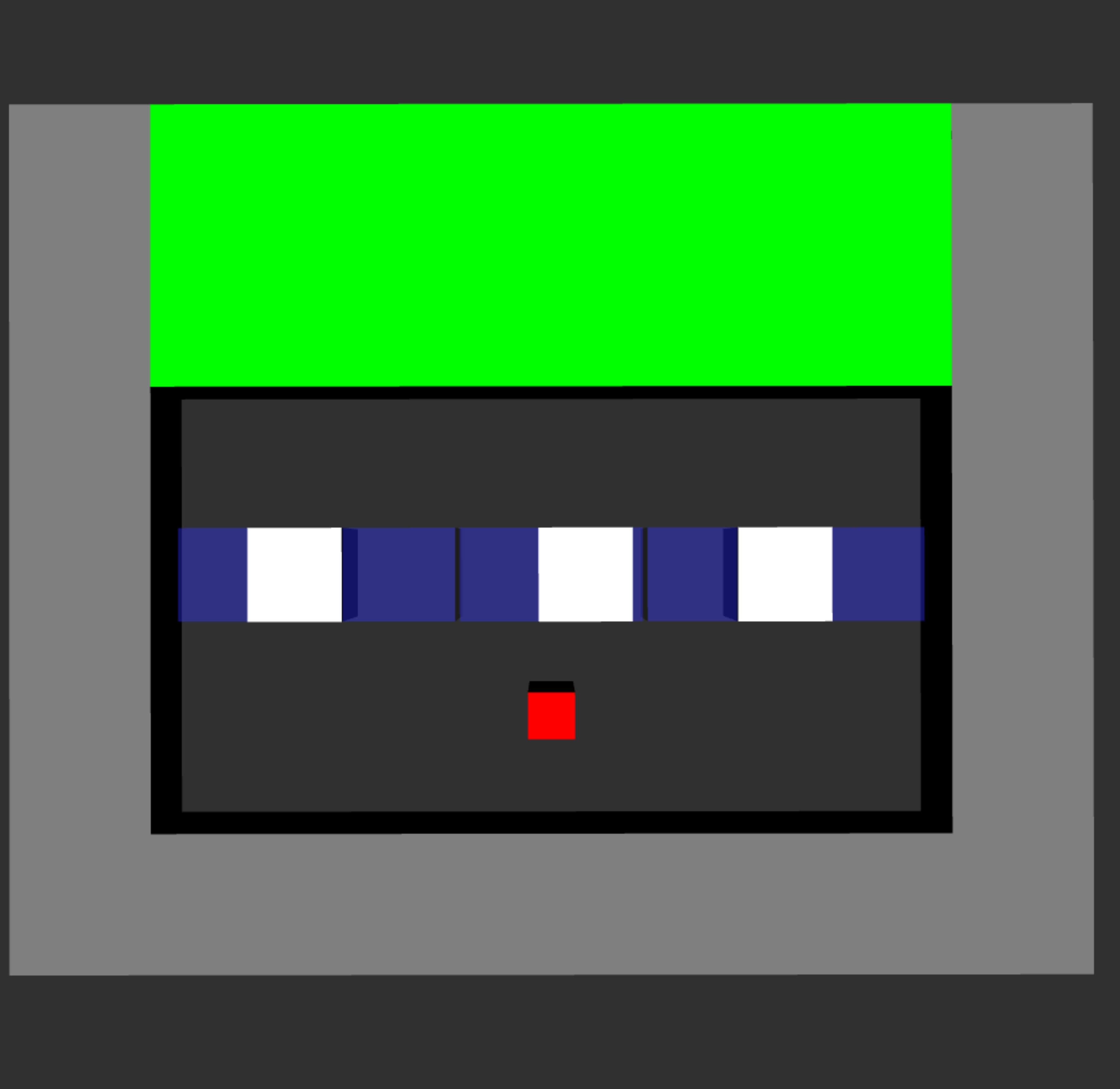} &
\includegraphics[width=0.2\textwidth]{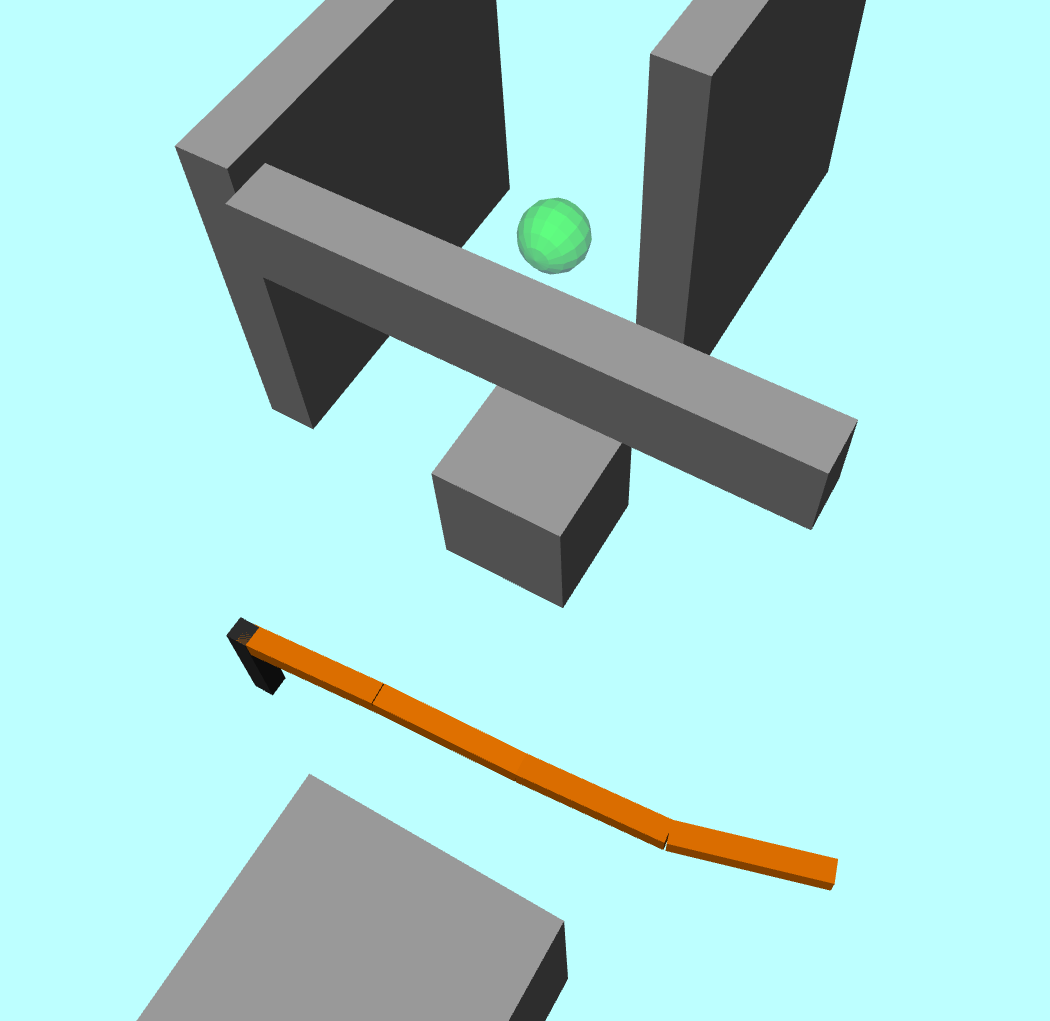} \\
(a) & (b)
\end{tabular}
\caption{The Passage scenario (a) and the 4DOF-Factory scenario (b).}
\label{f:problemScenarios}
\vspace{-0.5cm}
\end{figure}
 
A torque-controlled manipulator with 4 revolute joints must move from an initial state to a state where the end-effector lies inside a goal area (colored green in \fref{f:problemScenarios}(b)), without colliding with any of the obstacles. The state space is the joint product of joint-angles and joint-velocities. The control inputs are the joint-torques. The action space is set to be the maximum and minimum possible joint torque, resulting in 16 discrete actions affected by control errors. The dynamics of the manipulators are simulated by the ODE physics engine\ccite{Smi07:ODE} with integration step size $\delta t=0.004s$ (in\ccite{hoergerISRR2019} we studied a variant of this problem with a significantly more expensive transition function where we used $\delta t=0.0001s$). The observations are noisy too and consist of the position of the end-effector inside the robot's workspace and joint velocities, resulting in a 7-dimensional continuous observation space. The initial state is known exactly, for which the joint angles and velocities are zero. The robot enters a terminal state and receives a reward of 1,000 upon reaching the goal. To encourage the robot to reach the goal area quickly, it receives a small penalty of -1 for every other action. Collision result in a terminal state too and are penalised by -500. The discount factor is 0.98.

\begin{figure*}[h]
\centering
\small
\begin{tabular}{c@{\hskip0pt}c@{\hskip0pt}c@{\hskip0pt}c}
\includegraphics[width=0.3\textwidth]{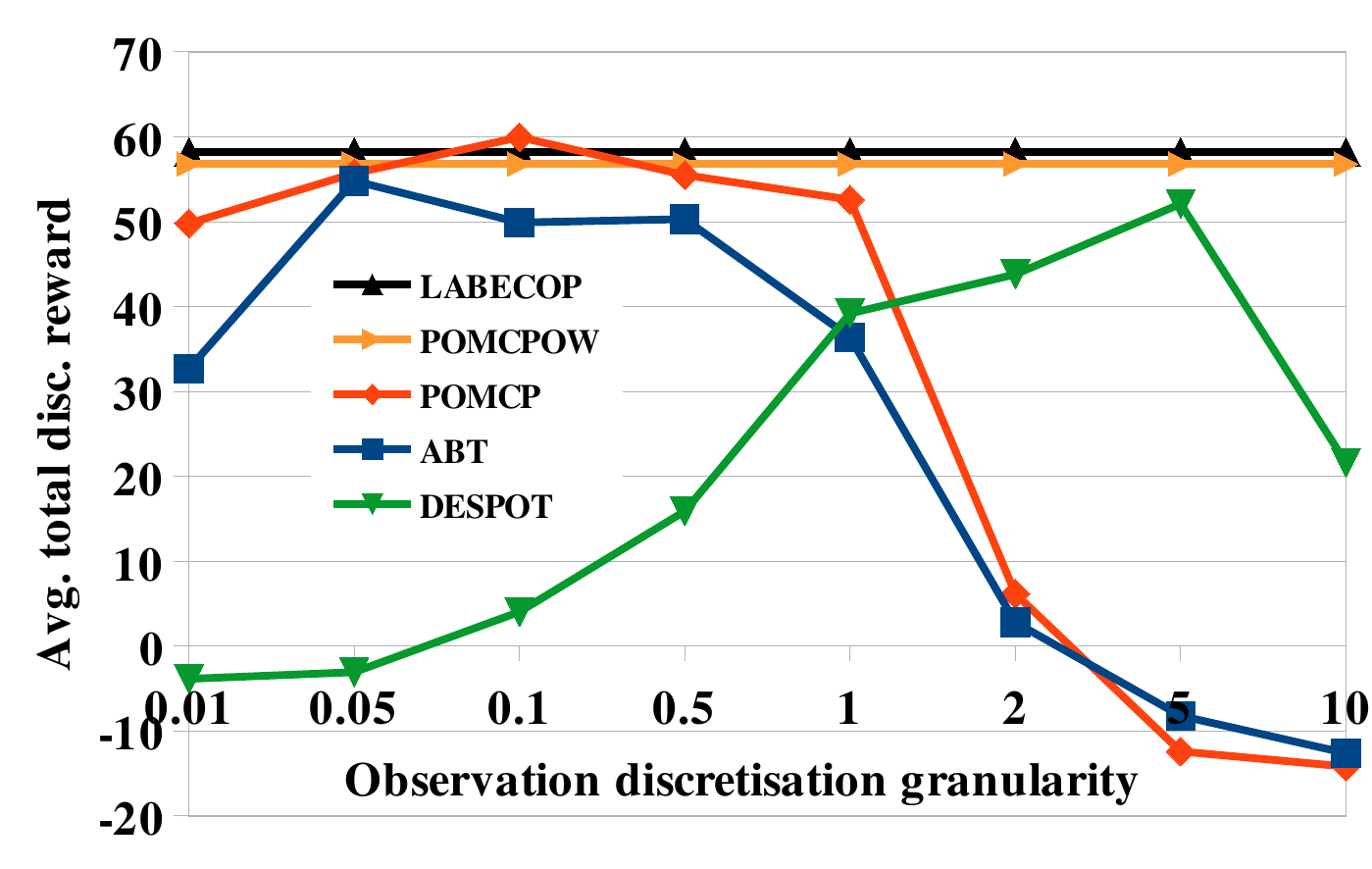} &
\includegraphics[width=0.3\textwidth]{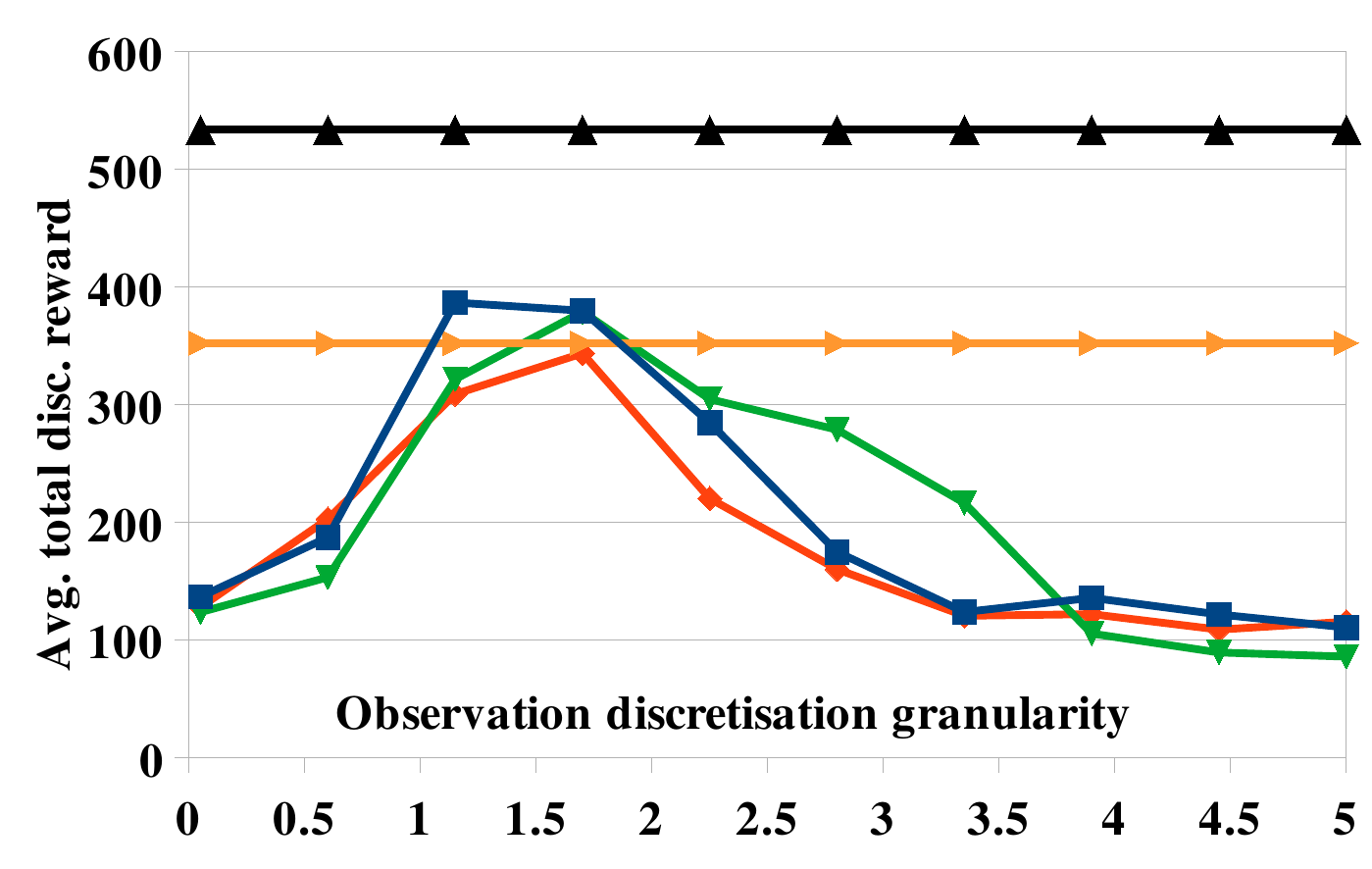} &
\includegraphics[width=0.3\textwidth]{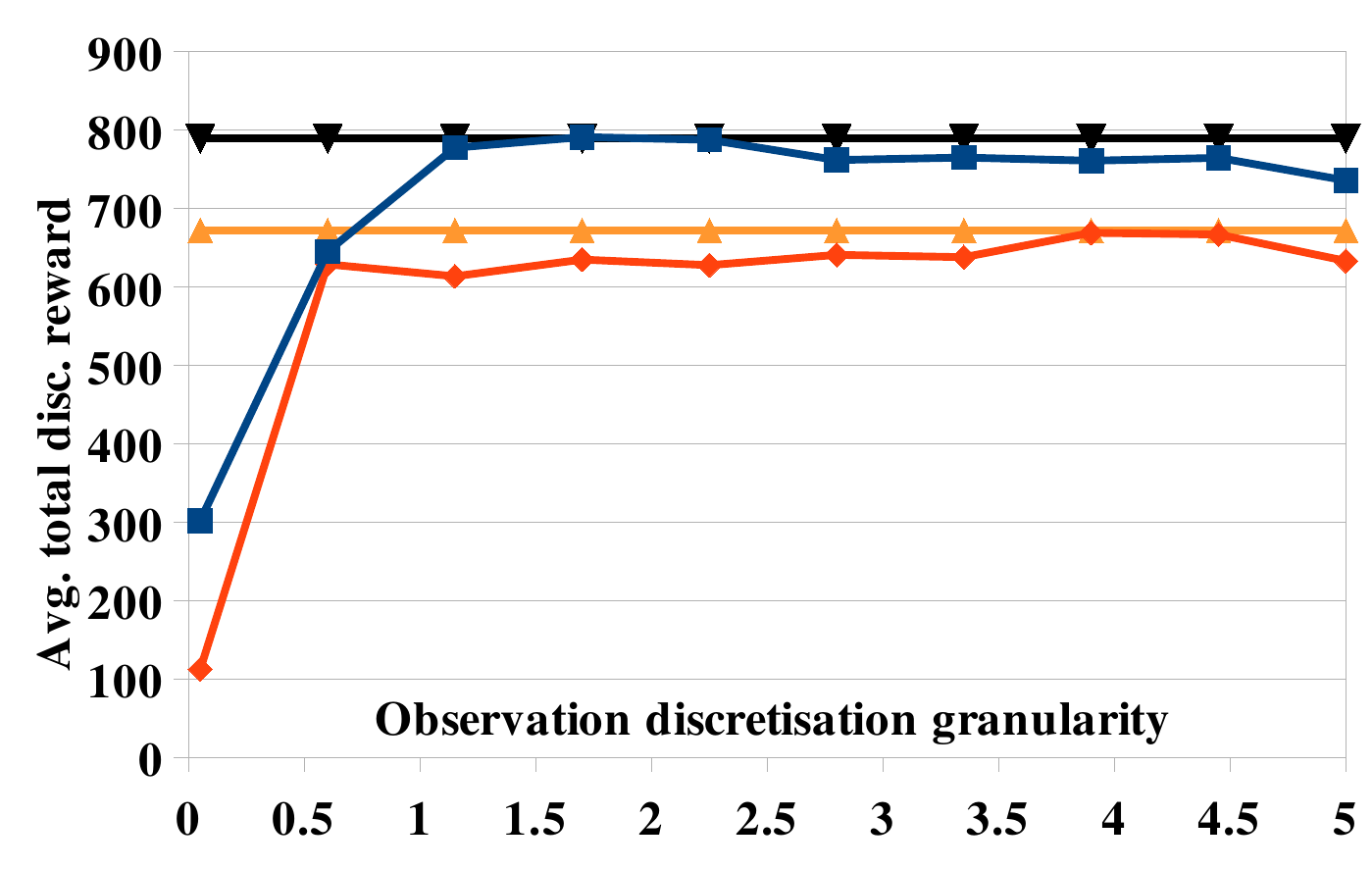} \\
(a) & (b) & (c)
\end{tabular}
\caption{Average total discounted reward of \solverShort, POMCPOW, POMCP, ABT and DESPOT on the LightDark1D (a), Passage (b) and 4DOF-Factory (c) scenarios. The $x$-axis represents the discretisation granularity of the observation space (from fine to coarse), whereas the $y$-axis represents the average total discounted reward. Note that \solverShort and POMCPOW don't discretise the observation space. Therefore their results are included as horizontal lines.\vspace{-0.5cm}}
\label{f:plots}
\end{figure*} 

\vspace{-0.2cm}
\subsection{Experimental setup} 
All three problem scenarios and the solvers were implemented in C++ using the OPPT-framework\ccite{hoerger2018software}. For ABT we used the implementation provided by OPPT, whereas for DESPOT we used the code released by the original authors (\url{https://github.com/AdaCompNUS/despot}). Since the version of POMCPOW released by the authors is written in the Julia programming language, which is incompatible with OPPT, we re-implemented POMCPOW in C++. However, our implementation of POMCPOW achieved better results in the LightDark1D problem than reported in\ccite{sunberg2018online}. For POMCP we used our own implementation within OPPT.

All simulations were run single-threaded on an Intel Xeon Silver 4110 CPU with 2.1Ghz and 128GB of memory. We used a fixed planning time of 1s for each solver for the LightDark1D and Passage problems and 2s for the 4DOF-Factory problem. Note that POMCP, ABT and DESPOT require a discretised observation space. For the LightDark1D problem we used a grid-based discretisation whereas for the Passage and 4DOF-Factory problems we used a distance based discretisation, \ie two observations are considered equal if their euclidean distance is smaller than a given threshold. 

For all problems and solvers we first ran a set of systematic preliminary trials to determine the best parameters for each solver. The parameters and their respective values are shown in \tref{t:hyperparameters}. 

Despite out best efforts, we were unable to obtain results for DESPOT for the 4DOF-Factory problem. The reason is that DESPOTs strategy of expanding each belief using every action via the forward simulation of $K$ scenarios is computationally too demanding for this problem due to its large action space and expensive transition function. In our experiments it took, on average, $\sim$3s to expand a single belief using $K=50$ scenarios (a tenth of what is commonly used\ccite{somani2013despot}), which is already more than the allowed planning time of 2s.   

To study the effect of discretising the observation space on the performance of POMCP, ABT and DESPOT, we first ran a set of experiments in which we varied the discretisation granularity for each problem, \ie we varied the grid-size in the LightDark1D problem and the observation distance threshold in the Passage and 4DOF-Factory problems. We then ran a set of experiments using the best discretisation granularity for POMCP, ABT and DESPOT and compared the results with POMCPOW and \solverShort using 1,000 simulation runs. The results are shown and discussed in the next section.

\begin{table}[htb]\label{t:hyperparameters}
\centering
\vspace{-0.2cm}
\begin{tabular}{llll}
 & \textit{LightDark1D} & \textit{Passage} & \textit{4DOF-Factory} \\ \hline
\multicolumn{4}{l}{\solverShort}\\ 
$c$ & $20.0$ & $420.0$ & $10.0$ \\ \hline
\multicolumn{4}{l}{POMCPOW}\\ 
$c$ & $80.0$ & $500.0$ & $17.0$ \\
$k_o$ & $3.25$ & $15.25$ & $1.0$ \\
$\alpha_o$ & $0.01$ & $0.01$ & $0.01$ \\ \hline
\multicolumn{4}{l}{POMCP}\\ 
$c$ & $40.0$ & $400.0$ & $20.0$ \\ \hline
\multicolumn{4}{l}{ABT}\\ 
$c$ & $10.0$ & $550.0$ & $20.0$ \\ \hline
\multicolumn{4}{l}{DESPOT}\\ 
numScenarios & $500$ & $300$ & $-$ \\
searchDepth & $10$ & $8$ & $-$ \\
\end{tabular}
\caption{Hyperparameters for each solver used in the experiments. $c$ is the UCB1 exploration constant. $k_\obs$ and $\alpha_\obs$ are the progressive widening parameters for POMCPOW.}
\label{t:results}
\vspace{-0.5cm}
\end{table}

\subsection{Results}

\begin{table}\label{t:avg_rewards}
\centering
\vspace{0.2cm}
\begin{tabular}{|l|l|l|l|}
\hline
 & \textit{LightDark1D} & \textit{Passage} & \textit{4DOF-Factory} \\ \hline \hline
\solverShort & $58.2 \pm 1.1$ & $\mathbf{533.6 \pm 17.2}$ & $789.4 \pm 17.4$ \\ \hline
POMCPOW & $56.8 \pm 1.1$ & $352.3 \pm 40.4$ & $672.4 \pm 25.2$ \\ \hline
POMCP & $\mathbf{59.6 \pm 1.1}$ & $343.6 \pm 36.7$ & $669.0 \pm 19.7$ \\ \hline
ABT & $54.8 \pm 1.4$ & $386.8 \pm 39.9$ & $\mathbf{791.2 \pm 18.4}$ \\ \hline
DESPOT & $52.1 \pm 1.7$ & $379.0 \pm 38.4$ & $ - $ \\ \hline
\end{tabular}
\caption{Average total discounted reward and 95\% confidence intervals achieved \solverShort, POMCPOW, POMCP, ABT and DESPOT in the three problem scenarios. The average is taken over 1,000 simulation runs for each solver and scenario. The best result for each solver and problem is highlighted in bold.\vspace{-0.5cm}}
\label{t:results}
\vspace{-0.5cm}
\end{table}

\fref{f:plots} shows the average total discounted reward achieved by each solver in all tree test scenarios as we vary the discretisation granularity of the observation space. Since \solverShort and POMCPOW don't discretise the observation space, their results are added as horizontal lines. \tref{t:avg_rewards} shows the average total discounted rewards for POMCPOW and \solverShort as well as for POMCP, ABT and DESPOT where we used the best observation space discretisation granularities for the latter three solvers. 

Looking at the results for the LightDark1D and 4DOF-Factory problems, it might seem surprising that solvers (POMCP for LightDark1D and ABT for 4DOF-Factory) that rely on discretising the observations space outperform POMCPOW and slightly outperform \solverShort. LightDark1D has a one-dimensional observation space that is fairly easy to discretise, although the discretisation window in which POMCP outperforms the other methods is quite narrow as seen in \fref{f:plots}(a). For 4DOF-Factory the results in \fref{f:plots}(c) indicate that discretising the observation space causes no issues for POMCP and ABT despite the 7-dimensional continuous observation space, as long as the granularity is not too fine. The reason is that in this problem, it is paramount that the solver has a lookahead of at least 6 steps in order to compute a strategy that avoids collisions with the obstacles. Therefore finer discretisation granularities of the observation space hurt POMCP and ABT due to the increased branching factor of the lookahead trees. Additionally, the transition and observation errors in this problem are small, leading to many sampled observations and their resulting beliefs to be very similar. This enables POMCP and ABT to aggregate observations into 1-3 observation edges at most without a significant performance penalty. 

Although in both scenarios the observation space is rather easy to discretise, \solverShort still performs comparable to the best solver in each respective scenario. This indicates that the performance penalty of considering \textit{every} sampled observation for \solverShort is small, even for problems where this is not necessary.

In the Passage problem \solverShort clearly outperforms the other methods as seen in \fref{f:plots}(b). Due to the large initial uncertainty regarding the locations of the obstacles, the robot must plan with respect to many possible observations produced by the \texttt{SCAN} action, each of them leading to beliefs with different potential passages between the obstacles. If too many beliefs are aggregated via a coarse observation space discretisation, it is unlikely that the robot discovers a strategy that works well for all the aggregated beliefs. If the discretisation is too fine, the branching factor of the lookahead trees used by POMCPOW, POMCP, ABT and DESPOT increases drastically. Since the robot requires multiple \texttt{SCAN} actions to find a safe passage to the goal area, the lookahead becomes too short to discover the long-term benefit of collecting observations. In both cases the robot avoids the \texttt{SCAN} action altogether. 

\solverShort mitigates both issues. By considering \textit{every} sampled observations during planning, \solverShort can plan with respect to many different sequences of beliefs. At the same time, \solverShort doesn't "branch" over observations, resulting in longer lookaheads. Both properties enable \solverShort to quickly discover the long-term benefit of multiple observation-gathering actions. As a result, \solverShort significantly outperforms the other methods in this scenario.

\section{CONCLUSIONS}
Despite tremendous progress in on-line POMDP planning, computing robust policies for problems with continuous observation spaces remains challenging. State-of-the-art solvers typically rely on discretising the observation space or limiting the number of observations that are considered during planning, which often impairs robustness. In this paper we present \solver (\solverShort), a new on-line POMDP solver designed to alleviate these shortcomings. \solverShort samples an maintains a set of episodes that is used to lazily extract beliefs and a corresponding action-selection strategy on-the-fly as it samples new episodes,  enabling \solverShort to consider \textit{every} sequence of beliefs it encounters during planning, without having to discretise the observation space or maintain a lookahead tree. This results in significant gains in robustness for problems that require reasoning about many different sequences of beliefs. Experiments on three test scenarios indicate that \solverShort performs similar or better than state-of-the-art on-line solvers for problems with continuous observation spaces.

\addtolength{\textheight}{-12cm}   







\bibliographystyle{IEEEtran}
\bibliography{pomdpRefs-short}

\end{document}